\documentclass[sigconf,natbib=false]{acmart}

\usepackage{booktabs}
\usepackage{multirow}
\usepackage{balance}

\AtBeginDocument{%
}

\copyrightyear{2023}
\acmYear{2023}
\setcopyright{rightsretained}
\acmConference[CIKM '23]{Proceedings of the 32nd ACM International Conference on Information and Knowledge Management}{October 21--25, 2023}{Birmingham, United Kingdom}
\acmBooktitle{Proceedings of the 32nd ACM International Conference on Information and Knowledge Management (CIKM '23), October 21--25, 2023, Birmingham, United Kingdom}\acmDOI{10.1145/3583780.3615125}
\acmISBN{979-8-4007-0124-5/23/10}

\copyrightyear{2023}
\acmYear{2023}
\setcopyright{rightsretained}
\acmConference[CIKM '23]{Proceedings of the 32nd ACM International Conference on Information and Knowledge Management}{October 21--25, 2023}{Birmingham, United Kingdom}
\acmBooktitle{Proceedings of the 32nd ACM International Conference on Information and Knowledge Management (CIKM '23), October 21--25, 2023, Birmingham, United Kingdom}\acmDOI{10.1145/3583780.3615125}
\acmISBN{979-8-4007-0124-5/23/10}

\RequirePackage[
  datamodel=acmdatamodel,
  style=acmnumeric,
  ]{biblatex}

\begin{document}

\title{\dataset: A Multi-View Similar Case Retrieval Dataset}

\author{Qingquan Li}
\authornote{Equal contribution. Listing order is random.}
\author{Yiran Hu}
\authornotemark[1]
\email{{liqq20, huyr21}@mails.tsinghua.edu.cn}
\affiliation{%
  \institution{Tsinghua University}
  \city{Beijing}
  \country{China}
  \postcode{100089}
}

\author{Feng Yao}
\email{yaof20@mails.tsinghua.edu.cn}
\affiliation{%
  \institution{Tsinghua University}
  \city{Beijing}
  \country{China}
  \postcode{100089}
}

\author{Chaojun Xiao}
\email{xiaocj20@mails.tsinghua.edu.cn}
\affiliation{%
  \institution{Tsinghua University}
  \city{Beijing}
  \country{China}
  \postcode{100089}
}

\author{Zhiyuan Liu}
\email{liuzy@tsinghua.edu.cn}
\authornote{Corresponding authors.}
\affiliation{%
  \institution{Tsinghua University}
  \city{Beijing}
  \country{China}
  \postcode{100089}
}

\author{Maosong Sun}
\email{sms@tsinghua.edu.cn}
\affiliation{%
  \institution{Tsinghua University}
  \city{Beijing}
  \country{China}
  \postcode{100089}
}

\author{Weixing Shen}
\email{wxshen@tsinghua.edu.cn}
\authornotemark[2]
\affiliation{%
  \institution{Tsinghua University}
  \city{Beijing}
  \country{China}
  \postcode{100089}
}

\renewcommand{\shortauthors}{Li, Hu et al.}
\newcommand{\dataset}{MUSER}

\begin{abstract}
  Similar case retrieval (SCR) is a representative legal AI application that plays a pivotal role in promoting judicial fairness. However, existing SCR datasets only focus on the fact description section when judging the similarity between cases, ignoring other valuable sections (e.g., the court's opinion) that can provide insightful reasoning process behind.
  Furthermore, the case similarities are typically measured solely by the textual semantics of the fact descriptions, which may fail to capture the full complexity of legal cases from the perspective of legal knowledge.
  In this work, we present \dataset, a similar case retrieval dataset based on multi-view similarity measurement and comprehensive legal element with sentence-level legal element annotations. 
  Specifically, we select three perspectives (legal fact, dispute focus, and law statutory) and build a comprehensive and structured label schema of legal elements for each of them, to enable accurate and knowledgeable evaluation of case similarities. 
  The constructed dataset originates from Chinese civil cases and contains 100 query cases and 4,024 candidate cases. 
  We implement several text classification algorithms for legal element prediction and various retrieval methods for retrieving similar cases on \dataset. The experimental results indicate that incorporating legal elements can benefit the performance of SCR models, but further efforts are still required to address the remaining challenges posed by \dataset. The source code and dataset are released at \url{https://github.com/THUlawtech/MUSER}.
\end{abstract}




\maketitle

\vspace{-1em}
\begin{figure}[ht]
  \centering
  \includegraphics[width=0.45\textwidth]{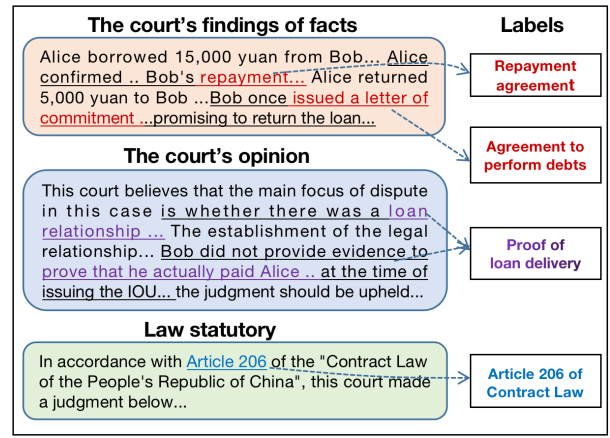} 
  \vspace{-1em}
  \caption{The valuable sections of a judgment document.} 
  \vspace{-1em}
  \label{img} 
\end{figure}
\vspace{-0.5em}

\section{Introduction}

Similar case retrieval (SCR) is a vital legal AI~\cite{li2020event, yao2023unsupervised, yao2022leven, zhong2020does} application that plays a strong role in achieving consistent judgment for maintaining judicial fairness. Given a case, the goal of the SCR task is to retrieve similar cases from the candidate pool according to the judgment criteria, and judges refer to similar cases to assist in the current case judgment.
Previous SCR works \cite{li2023thuir, li2023thuirr, liu2023investigating, liu2022query, ma2021retrieving, nigam2022nigam, shao2023intent, shao2020thuir, tran2019building, zhaowei2022legal, yu2022explainable} have achieved some success, \cite{kano2019coliee, xiao2019cail2019} measure the similarity of cases using textual information,  \cite{li2023sailer} develop similar case measurement strategies considering both subjective and objective evaluation, \cite{zhaowei2022legal} attend legal elements for SCR through contrastive learning. However, two main challenges of existing SCR datasets have constrained the progress of the case retrieval models:

\textbf{(1) Single-View.} Existing SCR datasets \cite{ma2021lecard, tran2019building} only analyze cases from the perspective of legal fact. 
However, as shown in Figure \ref{img}, in addition to the court's findings of fact, there are also sections such as ``court's opinion'' and ``law statutory'', which are very concerned about by legal experts and should not be ignored. \cite{trubek1980construction}
For example, As shown in Figure \ref{img}, the legal fact of the present case pertains to ``repayment agreement'', which is similar to one of the candidate cases.  
However, the dispute focus of the candidate case concerns ``limits on interest rates'', governed by ``Article 205 of the Contract Law'', which is different from the case in Figure \ref{img}. Although the legal fact suggests a high similarity, the two cases diverge in terms of their focus on different aspects of the repayment agreement, which leads to distinct dispute focuses and law statutory. Consequently, it is not appropriate to classify these two cases as strongly relevant.
The limitations of previous works have led to an incomplete measurement of case similarity. 
Therefore, it is particularly important to construct multi-view evaluation criteria for case similarity.

\textbf{(2) Lack of legal knowledge.} Existing SCR works \cite{buloubehavior, xiao2019cail2019} only judge similar cases through semantic similarity, which makes it difficult to consider deep-level legal knowledge in SCR. 
In judicial practice, cases are very complex. 
For example, the legal effects of loans between couples, loans between employer and employee, and loans between relatives are completely different, and cannot be inferred solely from the text.
Research in the law field \cite{alen47beyond} also attempts to design legal-element-based SCR methods for specific case types.
Therefore, it is necessary to consider structural legal knowledge when retrieving similar cases.

To alleviate the above issues, we present MUSER, a similar case retrieval dataset based on multi-view similarity measurement with sentence-level legal element annotations. 
We highlight MUSER with the following advantages:

\textbf{(1) Multi-View.} 
To address the issue that current SCR works mainly focus on fact description, we propose dividing judgment information into three dimensions: legal fact, dispute focus, and law statutory. Each dimension provides a summary and description of the case from a unique perspective. Integrating multi-view information into SCR can comprehensively summarize all the information of the case and meet a wide range of SCR needs, including those of ordinary people, lawyers, and judges. Our hypothesis is confirmed by the Supreme People's Court's issuance of guidance documents that divide the definition of similar cases into three dimensions.\footnote{\url{https://www.chinacourt.org/law/detail/2020/07/id/150187.shtml}}

\textbf{(2) Fine-grained legal element label schema.} As Figure \ref{img} shows, we propose a sentence-level legal element label system for the legal fact and dispute focus to represent the deep legal knowledge implied in cases. Our legal expert team presents a three-level label schema, which contains 22 1st-level labels, 190 2nd-level labels, and 505 3rd-level labels. We annotated labels at the sentence level for both query and candidate cases. Following manual labeling, we trained a deep neural model to detect legal elements from cases. 
Moreover, to our knowledge, previous SCR datasets mainly focus on criminal cases, whereas civil cases are more numerous and complex, necessitating an SCR dataset in this domain. In this paper, we propose MUSER, an SCR dataset based on multi-view similarity measurement with sentence-level legal element annotations.
Our dataset contains 100 query cases, each query has 100 candidates.  
All the query and candidate cases in MUSER are adopted from Chinese Civil Law, published by the Supreme People's Court of China.

We implement several legal element prediction models and SCR models to explore the challenges of MUSER. 
Meanwhile, we use our legal element labels to design an SCR model. 
Experimental results show that the performance of our model is better than those baseline models, which provides our introduction of multiple views and legal knowledge in SCR is effective.

\section{Dataset Construction}

In this section, we introduce our dataset construction methods. Our goal is to build a multi-view similar case retrieval dataset with sentence-level legal element annotation. Therefore, our task is to define a label schema, select the query cases and develop relevant judgment criteria. 
In this paper, we focus on private lending cases, as they are the most complex and voluminous among all civil cases.

\subsection{Label Schema Construction}

To comprehensively construct the legal element schema that describes private lending cases, we propose two types of labels: legal fact and dispute focus.
For each type of label, our legal expert team designs a large-scale, hierarchical three-level label schema, covering the legal elements that may appear in the vast majority of private lending cases.
These legal elements mainly involve private lending contracts, private lending relationships, private lending amounts, private lending litigation, etc.
Among them, the legal fact labels have 11 1st-level categories, 69 2nd-level categories, and 211 3rd-level categories;
the dispute focus labels have 11 1st-level categories, 121 2nd-level categories, and 294 3rd-level categories.
All labels are annotated at the sentence level. 
The legal fact labels are annotated in the ``court's findings of fact'' section of the judgment document, while the dispute focus labels are annotated in the ``court's opinions'' section.
Details can be found in our github link.

\subsection{Query and Candidate Selection}

Following \cite{ma2021lecard}, we collect 100 query cases from 7000 cases published by the Supreme People’s Court of China. Considering the diversity of label coverage, we sort the cases based on the number and types of labels and conduct uniform sampling. Since each case may cover multiple labels, we prioritize sampling cases with more label types in order to obtain a query set that covers more categories of labels.

For the candidate selection, we perform three strategies in sequence:
(1) Take the top 30 cases based on the cosine similarity of the text.
(2) Rank the document based on TF-IDF \cite{salton1988term} and BM25 \cite{robertson1995okapi}, and select the top 70 cases that appear in both rankings.
(3) If the candidate has less than 100 cases at this point, the leftover cases will be selected as the average of the above two rankings.

\subsection{Relevant Judgment Criteria and Annotation}

\begin{table}[]
\caption{Case relevance criteria on different dimensions.}
\vspace{-0.5em}
\label{tab:rel}
\resizebox{.98\linewidth}{!}{
\begin{tabular}{@{}cccc@{}}
\toprule
\textbf{Dimension}                                                       & \textbf{Description}                                                                                        & \textbf{Relevance}                                         & \textbf{Score} \\ \midrule
\multirow{7}{*}{\begin{tabular}[c]{@{}c@{}}Legal\\ fact\end{tabular}}    & \begin{tabular}[c]{@{}c@{}}Key facts are relevant, \\ general facts are relevant.\end{tabular}                     & \begin{tabular}[c]{@{}c@{}}strong\\ relevance\end{tabular} & 3              \\
                                                                         & \begin{tabular}[c]{@{}c@{}}Key facts are relevant \\ general facts are irrelevant.\end{tabular}                   & relevance                                                  & 2              \\
                                                                         & \begin{tabular}[c]{@{}c@{}}Key facts are irrelevant, \\ general facts are relevant.\end{tabular}                   & \begin{tabular}[c]{@{}c@{}}weak\\ relevance\end{tabular}   & 1              \\
                                                                         & \begin{tabular}[c]{@{}c@{}}Key facts are irrelevant, \\ general facts are relevant.\end{tabular}                   & irrelevance                                                & 0              \\ \midrule
\multirow{4}{*}{\begin{tabular}[c]{@{}c@{}}Dispute\\ focus\end{tabular}} & Identical key dispute focuses exist.                                                                         & \begin{tabular}[c]{@{}c@{}}strong\\ relevance\end{tabular} & 3              \\
                                                                         & \begin{tabular}[c]{@{}c@{}}Identical dispute focuses exist, \\ but not key.\end{tabular}                     & \begin{tabular}[c]{@{}c@{}}weak\\ relevance\end{tabular}   & 1              \\
                                                                         & No identical dispute focuses.                                                                               & irrelevance                                                & 0              \\ \midrule
\multirow{4}{*}{\begin{tabular}[c]{@{}c@{}}Law\\ statutory\end{tabular}} & \begin{tabular}[c]{@{}c@{}}Identical articles are cited \\ and have an impact on the judgement.\end{tabular} & relevance                                                  & 2              \\
                                                                         & \begin{tabular}[c]{@{}c@{}}Identical articles are cited \\ but have no impact on the judgement.\end{tabular} & irrelevance                                                & 0              \\
                                                                         & No identical articles are cited.                                                                             & irrelevance                                                & 0              \\ \bottomrule
\end{tabular}}
\vspace{-1em}
\end{table}

The ``multi-view'' similarity between query and candidate cases is judged by the following three dimensions: legal fact, dispute focus, and law statutory.

For legal facts, we define two types: key facts and general facts.
Key facts are facts that involve the plaintiff's requested rights, determine whether the case is a private lending case, and are typical in private lending cases.
General facts are usually incidental facts other than key facts, such as the calculation of interest.
For dispute focus, we also define two types: key dispute focus and general dispute focus. The key disputed focus includes the disputed focus generated around the plaintiff's core claims and the key facts of the case. The difference between key focus and general focus is whether they influence the plaintiff's basic claims and whether they involve the establishment of private lending. 
For the law statutory, the relevance judgment criteria are whether the same articles are cited and whether they have an impact on the judgment.
Due to the large number of articles, our legal expert team selects 35 common key articles in private lending cases as the relevant judgment criteria.
Table \ref{tab:rel} shows in detail the case relevance judgment criteria on different dimensions.\footnote{As the court's determination of the legal characters of a case, the disputes focus is a very critical factor in making a judgment. Therefore, we hold that as long as the key disputes focuses of two cases overlap, their similarity in this dimension is a strong relevance, regardless of whether general disputes focuses overlap.}
The candidates for each query will be sorted in descending order based on the sum of the three dimension relevance scores.
Candidates with a relevance score of no less than 7 are considered relevant to the query.

Legal elements label and case relevance annotation experts are students in civil law from globally renowned law schools.
They received a three-day training and performed trail annotation before the formal annotation.
During annotation, we grouped the annotation experts, and the results of each group were regularly checked by the group leader.
For case relevance annotation, we only reveal the legal element labels of the query case and hide them of the candidate cases, in order to help annotation experts better understand the query case and avoid the influence of legal element labels on the case relevance judgment.

\section{Data Analysis}

In this section, we aim to provide a deep understanding of MUSER through data analysis.

\subsection{Data Size}

\noindent{\textbf{Statistics.}}\quad    
Table \ref{tab:stats} shows the statistics of our dataset.
We consider candidate cases with relevance scores to the query cases that are not less than 7 as relevant cases for the query.

\begin{table}[t]
\caption{Dataset statistics of MUSER.}
\vspace{-0.8em}
\label{tab:stats}
\small
\begin{tabular}{@{}cc@{}}
\toprule
\textbf{Statistic}            & \textbf{Number} \\ \midrule
Total documents               & 4,024            \\
Total queries                  & 100             \\
Candidate cases per query     & 100             \\
Avg. relevant cases per query & 10.38           \\ \bottomrule
\end{tabular}
\vspace{-0.5em}
\end{table}

\noindent{\textbf{Text Size.}}\quad
Table \ref{tab:text-len} shows the text size of MUSER, including the average text length, average token number, and average sentence number.
For token statistics, we utilize jieba\footnote{\url{https://github.com/fxsjy/jieba}} for tokenization.
It is obvious that civil cases in MUSER are particularly long, which poses challenges for SCR, including extracting important information from long texts and encoding long texts with deep neural models.

\begin{table}[h]
\vspace{-0.3em}
\caption{Statistics of text size of MUSER. ``Avg. Len.'' is the average number of Chinese characters in the text, ``Avg. \#Token'' is the average number of words in the tokenized text.}
\vspace{-0.8em}
\label{tab:text-len}
 \resizebox{.98\linewidth}{!}{
\begin{tabular}{@{}cccc@{}}
\toprule
\textbf{Document Section}                                                     & \textbf{Avg. Len.} & \textbf{Avg. \#Token} & \textbf{Avg. \#Sentence} \\ \midrule
\begin{tabular}[c]{@{}c@{}}The court's findings\\ of fact\end{tabular} & 1,369.71              & 790.92                     & 28.07                         \\
The court's opinions                                                    & 1,629.34              & 900.94                     & 27.89                         \\
Total                                                                  & 2,999.05              & 1,691.86                    & 55.96                         \\ \bottomrule
\end{tabular}}
\vspace{-0.3em}
\end{table}

\noindent{\textbf{Legal Elements.}}\quad
Table \ref{tab:legal-element} shows detailed statistics of legal element annotations of MUSER.
Statistical results indicate the imbalance of our dataset,
which poses challenges for legal element prediction.

\begin{table}[h]
\vspace{-0.3em}
\caption{Statistics of legal element annotation of MUSER.}
\vspace{-0.8em}
\label{tab:legal-element}
\resizebox{.98\linewidth}{!}{
\begin{tabular}{@{}ccccc@{}}
\toprule
\textbf{Label Type} & \textbf{Document Section}                                                     & \textbf{\#Sentence} & \textbf{\#Label} & \textbf{\#Negative.} \\ \midrule
Legal fact    & \begin{tabular}[c]{@{}c@{}}The court's findings\\ of fact\end{tabular} & 112,951             & 40,636           & 78,916               \\
Dispute focus       & The court's opinions                                                    & 112,229             & 24,501           & 91,930               \\ \bottomrule
\end{tabular}}
\vspace{-0.3em}
\end{table}

\subsection{Data Distribution}

MUSER has data imbalance issues in legal element label distribution.
For the 1st-level labels, 3/11 of legal fact labels and 2/11 of dispute focus labels account for more than 50\% of the total instances.
However, most legal elements still have sufficient instances, with more than 1000 instances for 8/11 of legal fact labels and 6/11 of disputed focus labels.
Additionally, in actual judicial practice, the occurrence of different legal facts and disputed focuses follows a long-tail distribution, which proves that MUSER can serve as a real-world SCR dataset.
Therefore, inspired by \cite{wang2020maven}, we do not perform data augmentation or balancing during dataset construction.

\section{Experiment}

\subsection{Legal Element Prediction}

We evaluated several baseline models for predicting legal elements. We fine-tune the neural network models, including BERT\cite{devlin2019bert} and Lawformer\cite{xiao2021lawformer}, as a multi-label classification task. Our approach involves encoding the sentences using these deep neural network models and utilizing a fully-connected layer for classification. To create the train and test sets, we randomly sample 80\% of the sentences for training and keep the remaining as the test set.
We do not consider the hierarchy of our label schema and directly used the 3rd-level legal elements as the classification labels.
The results of the legal element prediction are shown in Table\ref{tab:labelprediction-results}.
The experimental results show that due to the complexity of the label schema and data imbalance, legal element classification is a challenging task.
In addition, how to incorporate the hierarchical information between legal elements is a key focus of future work.

\begin{table}[]
\caption{Evaluation of the legal element prediction task.}
\vspace{-0.8em}
\label{tab:labelprediction-results}
\small
\begin{tabular}{@{}ccccc@{}}
\toprule
\textbf{Label Type}                     & \textbf{Model}     & \textbf{Precision} & \textbf{Recall} & \textbf{F1}    \\ \midrule
\multirow{2}{*}{\textbf{Legal fact}}    & \textbf{BERT}      & \textbf{64.07}     & 56.43           & 60.01          \\
                                        & \textbf{Lawformer} & 63.56              & \textbf{59.65}  & \textbf{63.10} \\ \midrule
\multirow{2}{*}{\textbf{Dispute focus}} & \textbf{BERT}      & \textbf{54.70}     & 41.50           & 46.86          \\
                                        & \textbf{Lawformer} & 53.25              & \textbf{44.62}  & \textbf{48.83} \\ \bottomrule
\end{tabular}
\vspace{-1em}
\end{table}

\subsection{Similar Case Retrieval}

\noindent{\textbf{Baseline Models.}\quad
We evaluate several competitive retrieval models, which are widely used in the SCR task, on \dataset.
Three types of retrieval models are involved, including
(1) Bag-of-words (BoW) IR models.
Following \cite{ma2021lecard}, we select BM25 \cite{robertson1995okapi}, TF-IDF \cite{salton1988term}, and LMIR \cite{ponte1998language} as bag-of-words similar case retrievers.
(2) Deep neural models.
We consider Lawformer \cite{xiao2021lawformer}, a law-specific pre-trained language model capable of processing long texts, as the deep neural retrieval model.
Specifically, we fine-tune Lawformer with a text pair relevance classification task.
The task input contains the court's findings of fact and opinion of the query and candidate cases.
A \texttt{[CLS]} token is added to the beginning of the input, and two \texttt{[SEP]} tokens are used, one to separate the query-candidate pair, and the other added to the end of the input.
Global attention is added to \texttt{[CLS]} and query case.
The hidden state vector of \texttt{[CLS]} outputted by Lawformer is fed into a fully-connected layer for relevance classification.
Finally, the candidates are sorted in descending order based on the probability of being predicted as relevant to the query.
(3) SCR models based on legal elements.
Tradition IR models only consider textual information, lacking attention to professional legal concepts. 
Aiming at this issue, we present an SCR model based on our annotated legal elements including legal fact, dispute focus, and law statutory.
Specifically, labels of a certain level in a specific dimension of a case are represented as a vector $\mathcal{L} = (l_1, \dots, l_i, \dots, l_n)$, where $l_i$ is 1 when label $i$ is annotated, otherwise 0.
For label vectors of the same level and dimension in the query and candidate cases, we calculate their cosine similarity as the relevance score.
For the three dimensions, we compute the weighted sum of the relevance scores by the number of labels for each level.
Finally, the weighted sum of the relevance scores for three dimensions between query and candidate cases represents their final similarity score.
Assigning different weights to different dimensions can adjust the degree of attention to them during similar case retrieval.

\noindent{\textbf{Settings.}}\quad
We test BoW models and the legal-element-based model on the overall queries.
To compare the deep neural model with other retrieval models, we randomly sample 80\% of the queries as the train set for Lawformer, and test all baselines on the rest 20\% queries as the test set.
We evaluate baseline models on precision metrics including P@5, P@10, and Mean Average Precision (MAP), and ranking metrics including NDCG@10, NDCG@20, and NDCG@30.

For baseline implementation, we use gensim\footnote{\url{https://radimrehurek.com/gensim/}} to implement BM25, TF-IDF, and LMIR, and set all parameters to default values.
We fine-tuned Lawformer on 8 Nvidia RTX 2080Ti GPUs.
Due to memory limitations, we set the maximum length of both query and candidate text to 600.
For the legal-element-based model, we set the weights of legal fact, dispute focus, and law statutory to 0.5, 0.4, and 0.1, to attend more to the first two dimensions.

\begin{table}[]
\caption{Evaluation of retrieval models on different query sets of MUSER. ``N@(10/20/30)'' is NDCG@(10/20/30), ``LFM'' is Lawformer, ``LE'' is our proposed legal-element-based model.}
\vspace{-0.8em}
\label{tab:retrieval-results}
\resizebox{.98\linewidth}{!}{
\begin{tabular}{@{}c|c|cccccc@{}}
\toprule
                                            & \textbf{Model}    & \textbf{P@5}   & \textbf{P@10}  & \textbf{MAP}   & \textbf{N@10} & \textbf{N@20} & \textbf{N@30} \\ \midrule
\multirow{4}{*}{\textbf{\begin{tabular}[c]{@{}c@{}}Overall\\ Query Set\end{tabular}}} & \textbf{BM25}     & 63.60          & 48.60          & 79.24          & 23.68            & 21.98            & 20.53            \\
                                            & \textbf{TF-IDF}   & 72.20          & 59.80          & 81.52          & 23.96            & 22.35            & 21.47            \\
                                            & \textbf{LMIR}     & 68.00          & 53.70          & \textbf{84.40} & 26.33            & 23.54            & 21.89            \\ \cmidrule(l){2-8} 
                                            & \textbf{LE}     & \textbf{77.20} & \textbf{65.50} & 83.23          & \textbf{28.96}   & \textbf{26.02}   & \textbf{24.51} \\ \midrule
\multirow{5}{*}{\textbf{Test Set}}          & \textbf{BM25}     & 78.00          & 53.50          & 91.76          & 21.80            & 19.54            & 17.48            \\
                                            & \textbf{TF-IDF}   & 80.00          & 63.50          & 85.23          & 20.61            & 18.30            & 17.85            \\
                                            & \textbf{LMIR}     & \textbf{83.00} & 63.50          & \textbf{92.55} & 28.57            & 24.43            & 22.04            \\
                                            & \textbf{LFM} & 28.00              & 17.50              & 65.00              & 3.83                & 4.01                & 3.93                \\ \cmidrule(l){2-8} 
                                            & \textbf{LE}     & 81.00          & \textbf{71.50} & 87.01          & \textbf{31.82} & \textbf{27.01} & \textbf{25.29} \\ \bottomrule
\end{tabular}}
\vspace{-1em}
\end{table}

\noindent{\textbf{Results.}}\quad
The similar case retrieval results are shown in Table \ref{tab:retrieval-results}.
We can observe that
(1) Our proposed legal-element-based SCR model can outperform other baselines significantly, especially achieving comprehensive superiority in ranking metrics.
We attribute this to our fine-grained annotated labels that can make the retrieval model focus on the key legal elements in the case, while other retrieval models only pay attention to the textual information.
(2) The performance of the deep neural model is comparatively poor.
We suspect that the possible reason is the truncation of case text causes the model to miss important information.
(3) Overall, the civil case retrieval task is challenging, especially in terms of ranking metrics.
Compared to criminal cases, civil cases have more complex legal relationships and diverse legal elements, making similar civil case retrieval more difficult.
Future research is needed to explore SCR models specifically designed for civil cases.

\section{Conclusion}

In this paper, we propose MUSER, a similar case retrieval dataset for Chinese civil law systems.
Compared with other SCR datasets that only consider legal facts, we propose a multi-view definition of similar cases that incorporates dispute focuses and law statutory.
To incorporate legal knowledge into SCR, we design a large-scale label schema to represent the legal elements in the cases and conducted sentence-level annotation.
The experimental results show the challenge in legal element prediction and civil SCR tasks, providing direction for future work.

\begin{acks}
This work is supported by the National Key Research and Development Program of China (No. 2022YFC3301500), the National Natural Science Foundation of China (No. 62236004), Institute for Artificial Intelligence and Law, Tsinghua University, Institute Guo Qiang at Tsinghua University, and Jiangsu Collaborative Innovation Center for Language Ability, Jiangsu Normal University, Xuzhou, China.
\end{acks}

\newpage
\balance

\printbibliography


\end{document}